\documentclass{article}

\PassOptionsToPackage{numbers, compress}{natbib}

\usepackage[preprint]{neurips_2025}

\usepackage[utf8]{inputenc} 
\usepackage[T1]{fontenc}    
\usepackage{hyperref}       
\usepackage{url}            
\usepackage{booktabs}       
\usepackage{amsfonts}       
\usepackage{nicefrac}       
\usepackage{microtype}      
\usepackage{xcolor}         

\usepackage{amsmath}
\usepackage{graphicx}
\usepackage{enumitem}
\usepackage{multirow}
\usepackage{multicol}
\usepackage{rotating}
\usepackage{wrapfig}
\usepackage{caption}

\title{Event-Driven Dynamic Scene Depth Completion}

\author{Zhiqiang Yan$^{1}$ \quad Jianhao Jiao$^{2}$ \quad Zhengxue Wang$^{3}$ \quad Gim Hee Lee$^{1}$\\
$^1$NUS\quad
$^2$UCL\quad
$^3$NJUST\\
{\tt\small \{yanzq, gimhee.lee\}@nus.edu.sg}
}

\begin{document}

\maketitle

\begin{abstract}

Depth completion in dynamic scenes poses significant challenges due to rapid ego-motion and object motion, which can severely degrade the quality of input modalities such as RGB images and LiDAR measurements. 
Conventional RGB-D sensors often struggle to align precisely and capture reliable depth under such conditions. 
In contrast, event cameras with their high temporal resolution and sensitivity to motion at the pixel level provide complementary cues that are 
beneficial in dynamic environments.
To this end, we propose \textbf{EventDC}, the first event-driven depth completion framework. 
It consists of two key components: Event-Modulated Alignment (EMA) and Local Depth Filtering (LDF). 
Both modules adaptively learn the two fundamental components of convolution operations: offsets and weights conditioned on motion-sensitive event streams. 
In the encoder, EMA leverages events to modulate the sampling positions of RGB-D features to achieve pixel redistribution for improved alignment and fusion. 
In the decoder, LDF refines depth estimations around moving objects by learning motion-aware masks from events. 
Additionally, EventDC incorporates two loss terms to further benefit global alignment and enhance local depth recovery. 
Moreover, we establish the first benchmark for event-based depth completion comprising one real-world and two synthetic datasets to facilitate future research. Extensive experiments on this benchmark demonstrate the superiority of our EventDC. 
%

\end{abstract}

\section{Introduction}

\begin{figure}[t]
\centering
\includegraphics[width=0.94\columnwidth]{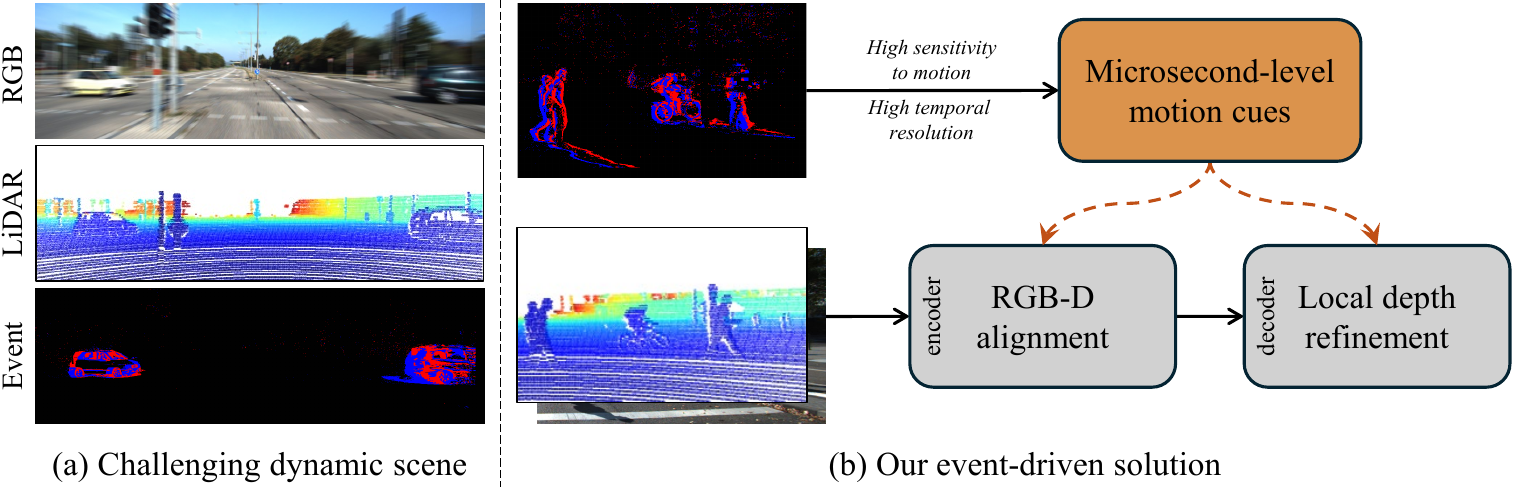}\\
\vspace{-3pt}
\caption{Data example and our solution for depth completion in dynamic environments. 
Leveraging high temporal resolution and motion sensitivity, event provides valuable complementary information for depth completion in dynamic scenes. 
Multiple event streams are aggregated for clear visualization. 
}
\vspace{-8pt}
\label{fig_data_solution}
\end{figure}

Depth completion \cite{Uhrig2017THREEDV,ma2018sparse,park2020nonlocal,tang2024bilateral,yan2024tri} aims to predict dense depth from sparse measurements, typically using auxiliary modalities such as RGB images. 
As a cornerstone of 3D perception, it plays a crucial role in a wide range of downstream applications including self-driving \cite{zhao2021adaptive,wang2024improving,liang2025distilling}, augmented reality \cite{song2020channel,yan2022learning,yan2024completion}, 
scene understanding \cite{wang2023lrru,shao2023nddepth,zuo2024ogni}, \textit{etc}. 
Although recent methods have demonstrated impressive results in static scenes, dynamic environments remain highly challenging. 
As illustrated in Fig.~\ref{fig_data_solution}(a), the rapid ego-motion results in blurry RGB images and misalignment with LiDAR measurements, while fast-moving objects further exacerbate depth inaccuracies in their vicinity. These challenges make precise depth completion even more difficult.

The unique characteristics of event cameras \cite{gallego2020event,gallego2018unifying,gehrig2024low} provide a compelling complement to conventional RGB-D sensors in dynamic scenes. 
Their microsecond-level temporal resolution enables the reliable capture of rapid ego-motion without introducing motion blur, and their asynchronous change-driven operation makes them inherently well-suited for detecting fast-moving objects. 
These properties help mitigate the limitations of RGB-D measurements by offering temporally consistent and low-latency signals, particularly in regions where traditional sensors often fail. 
As a result, event-based sensing proves especially advantageous for depth completion in highly dynamic environments.

In this work, we present EventDC, a novel depth completion framework that leverages event data to tackle the challenges posed by dynamic scenes. 
As shown in Fig.~\ref{fig_data_solution}(b), the core idea is to exploit the unique properties of event streams to guide depth completion especially in motion-affected regions. 
To this end, our EventDC incorporates two key components: Event-Modulated Alignment (EMA) and Local Depth Filtering (LDF). 
EMA is an encoder-side module that adaptively adjusts convolutional sampling positions using event information to achieve pixel redistribution for enhanced global alignment and more effective multi-modal fusion between RGB and LiDAR features. 
Furthermore, it incorporates a structure-aware loss to mitigate the RGB-D inconsistency caused by rapid ego-motion. 
LDF is a decoder-side module that focuses on refining depth around moving objects. It first learns motion masks from event streams to get the regions influenced by object motion. The learned masks are then used by LDF with a local motion-aware constraint to facilitate more accurate depth predictions in these regions. 
Concurrently, the two modules enable our EventDC to address both global misalignment and local depth inaccuracies for handling complex scenarios involving motion.

Additionally, depth completion based on event cameras remains an underexplored area with no existing event-based depth completion datasets to date.
To address this gap, we introduce the first event-based depth completion benchmark, which includes a real-world dataset \textit{EventDC-Real}, a semi-synthetic dataset \textit{EventDC-SemiSyn}, and a fully synthetic dataset \textit{EventDC-FullSyn}.

In summary, our contributions are as follows:

\begin{itemize}[leftmargin=2em,itemsep=2pt]
\item 
To the best of our knowledge, we are the first to introduce EventDC, a novel event-driven depth completion framework designed to address the challenges of dynamic environments. 
\item 
We present two event-driven modules: EMA and LDF which are designed to mitigate the global misalignment caused by ego-motion and local depth inaccuracies due to object motion. Additionally, these two modules are jointly supported by two dedicated loss constraints. 
\item 
To foster further research, we build the first event-based benchmark for depth completion. Extensive experiments across these datasets demonstrate the superiority of our approach, with up to 12.8\% improvement on the best-performing dataset over suboptimal methods. 
\end{itemize}

\section{Related Work}

\textbf{Depth Completion.}
Early depth completion methods \cite{Uhrig2017THREEDV,ma2018sparse,ku2018ipbasic,2020Confidence,vangansbeke2019,2020FromLu} focus on predicting dense depth maps directly from sparse inputs. For example, IP-Basic \cite{ku2018ipbasic} uses traditional image processing techniques to densify sparse depth without deep learning. In contrast, \citet{Uhrig2017THREEDV} introduce Sparsity Invariant CNNs, which adapt convolutional operations to varying input densities to ensure consistent performance. S2D \cite{ma2018sparse} employs an encoder-decoder architecture to progressively densify sparse depth input. FusionNet \cite{vangansbeke2019} integrates global context and local structures with a confidence-driven refinement mechanism. 
\citet{2020Confidence} present a confidence propagation method within CNNs to improve sparse depth regression by modeling uncertainty. Guided depth completion using color images has gained significant traction \cite{tang2020learning,zhao2021adaptive,yan2024tri,hu2020PENet,zhang2023cf,zhou2023bev,tang2024bilateral,yan2023desnet,yan2023learnable}. Dynamic filtering techniques \cite{tang2020learning,yan2022rignet,yan2023rignet++} generate adaptive filtering kernels from color images for effective extraction of depth features. Methods such as FuseNet \cite{chen2019learning}, PointDC \cite{yu2023aggregating}, BEVDC \cite{zhou2023bev}, and TPVD \cite{yan2024tri} further enhance depth completion by incorporating raw point clouds. Moreover, priors of the depth foundation models are used to improve generalization \cite{park2024depth,park2024simple,viola2024marigold,gregorek2024steeredmarigold}. Recently, SigNet \cite{yan2024completion} redefines depth completion as enhancement, densifying sparse depth with non-CNN methods, and then refines it through a degradation-aware framework. In addition, SPN techniques \cite{cspn_pami,Cheng2020CSPN++,park2020nonlocal,xu2020deformable,hu2020PENet,lin2023dyspn,yan2024tri}, which serve as effective refinement modules, can further enhance performance.

\textbf{Event-Based Depth Estimation.} 
Depth estimation with event cameras \cite{gallego2018unifying,ghosh2022event,gallego2020event,gallego2017event,gehrig2021combining,hidalgo2020learning,liu2024event,wei2024fusionportablev2} attracts growing interest due to the high temporal resolution, dynamic range, and low latency of asynchronous vision sensors. Early methods reconstruct depth solely from event streams such as the end-to-end framework by \citet{hidalgo2020learning}, the multi-view stereo pipeline EMVS \cite{rebecq2018emvs}, and unsupervised learning approaches for depth and egomotion \cite{zhu2019unsupervised}. DERD-Net \cite{hitzges2025derd} further exploits 3D convolutions and recurrence on event-based disparity space images, and Zhu et al. \cite{zihao2017event} propose a self-supervised framework for joint depth and optical flow estimation. Recent works leverage additional modalities to enhance event-based depth estimation. EMoDepth \cite{zhu2023self} temporally aligns events and intensity frames for self-supervised monocular depth learning. \citet{muglikar2021event} propose event-guided illumination control for active depth sensors. 
SRFNet \cite{pan2024srfnet} fuses frame and event features for fine-grained depth prediction with improved structure in both daytime and nighttime scenes. 
SDT \cite{zhang2024novel} combines spiking neural networks and transformers for efficient depth estimation. 
Furthermore, contrast maximization that emerges as a fundamental principle for event-based motion, depth, and optical flow estimation \cite{gallego2018unifying,shiba2024secrets} 
has inspired many subsequent works.

\textbf{Dynamic Convolution.} 
Dynamic convolution is a method that adjusts the convolution operation based on input features, and it gains significant attention in computer vision tasks. Techniques such as graph convolution and deformable convolution serve as specific manifestations of dynamic convolution. For example, ACMNet and GraphCSPN build graph structures to enable effective multi-modal fusion and refinement. 
STN \cite{jaderberg2015spatial} introduces the concept of spatially transforming features within a network, although training such a mechanism is a challenging task. Following this, DFN \cite{jia2016dynamic} proposes an approach that adapts filter parameters based on input features despite maintaining fixed kernel sizes. Deformable Convolution \cite{dai2017deformable,zhu2019deformable} takes a different approach with the focus on dynamically adjusting sampling locations by generating offsets based on the geometric properties of objects. 
Similarly, Active Convolution \cite{jeon2017active} improves sampling by adjusting the locations while keeping the kernel shape fixed. 
More recently, GuideNet \cite{tang2020learning} develops a guided convolution block specifically designed for multi-modal data. Despite these innovations, dynamic mechanisms often add considerable complexity. To address this issue, RigNet \cite{yan2022rignet} simplifies the dynamic guidance process by employing convolution factorization combined with attention \cite{hu2018squeeze}.

\section{Our Method}
\subsection{Background}
The core of dynamic convolution lies in the adaptive determination of sampling positions and weights. Graph Convolutional Networks (GCNs) \cite{kipf2016semi,velivckovic2017graph} and Deformable Convolutional Networks (DCNs) \cite{dai2017deformable, zhu2019deformable} serve as representative implementations of this concept. GCNs define sampling locations as neighboring nodes within the graph structure and compute adaptive weights during the aggregation stage. On the other hand, DCNs determine sampling locations through learned offsets and obtain adaptive weights by modulating predefined kernel weights with learned scalars. 
Both GCNs and DCNs can be viewed as extensions of standard convolutional operations, where the sampling locations and weights are made learnable and structure-aware. We use DCNs as an example to illustrate this dynamic learning process.

Specifically, DCNv1~\cite{dai2017deformable} introduces learnable offsets for each sampling location to shift adaptively. Subsequently, DCNv2~\cite{zhu2019deformable} further incorporates a learnable modulation scalar for each sampling position that enables the assignment of varying importance to different locations. 
Given an input feature map $\mathbf{x}$ and a convolutional kernel with $K$ sampling positions, let $\mathbf{w}_k$ and $\mathbf{p}_k$ denote the weight and the pre-defined offset of the $k$-th position, respectively. DCNv2 can be formulated as: 
\begin{equation}
\hat{\mathbf{x}}(\mathbf{p}_0) = \sum_{k=1}^{K} \mathbf{w}_k \cdot \mathbf{x}(\mathbf{p}_0 + \mathbf{p}_k + \Delta \mathbf{p}_k) \cdot \Delta \mathbf{m}_k,
\label{eq_dcn}
\end{equation}
where $\mathbf{p}_0$ denotes the reference location, and $\Delta \mathbf{p}_k$ and $\Delta \mathbf{m}_k$ are the learnable offset and modulation scalar, respectively. Note that $\mathbf{w}_k$ and $\Delta \mathbf{m}_k$ can be jointly interpreted as a unified learnable term. As a result, the adaptive adjustment of offset and weight in DCNv2 provides a foundation for leveraging event data to tackle the challenges posed by fast motion in depth completion.

\begin{figure}[t]
\centering
\includegraphics[width=0.98\columnwidth]{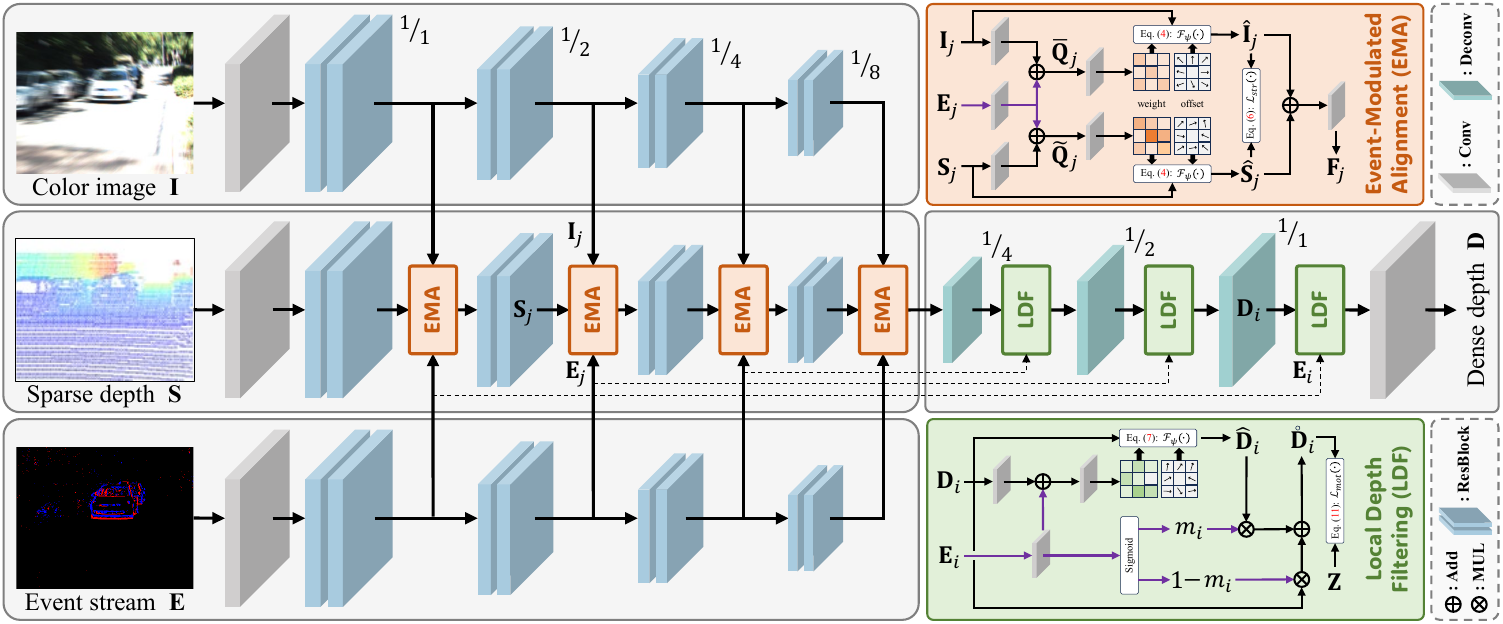}\\
\vspace{-3pt}
\caption{
Pipeline of our EventDC. The color image $\mathbf{I}$, sparse depth $\mathbf{S}$, and event stream $\mathbf{E}$ are first processed by three structurally identical encoders. At each stage, the Event-Modulated Alignment (EMA) block leverages event features to align and fuse RGB-D representations. In the decoder, the Local Depth Filtering (LDF) unit further enhances depth estimation around moving objects, guided by the inherent sensitivity of events to motion and reinforced by local motion-aware constraints. 
}
\label{fig_pipeline}
\vspace{-2pt}
\end{figure}

\subsection{EventDC Architecture}
\textbf{Overview.} 
In highly dynamic environments, the proposed approach is designed to alleviate the adverse effects of fast motion that include global misalignment and local depth inaccuracies caused by ego-motion and object motion, respectively. 
Fig.~\ref{fig_pipeline} illustrates the pipeline of our EventDC, which begins by employing three structurally consistent encoders to extract features from the color image $\mathbf{I}$, sparse depth $\mathbf{S}$, and event data $\mathbf{E}$. This yields multi-scale representations $\{ \mathbf{I}_1,\mathbf{I}_2,\mathbf{I}_3,\mathbf{I}_4 \}$, $\{ \mathbf{S}_1,\mathbf{S}_2,\mathbf{S}_3,\mathbf{S}_4 \}$, and $\{ \mathbf{E}_1,\mathbf{E}_2,\mathbf{E}_3,\mathbf{E}_4 \}$ at the $\{ ^{1}\!/\!_{1}, ^{1}\!/\!_{2}, ^{1}\!/\!_{4}, ^{1}\!/\!_{8} \}$ stages, respectively. 
In the decoder, three deconvolution layers are applied to progressively generate $\{ \mathbf{D}_3,\mathbf{D}_2,\mathbf{D}_1 \}$ at the $\{ ^{1}\!/\!_{4}, ^{1}\!/\!_{2}, ^{1}\!/\!_{1} \}$ stages, respectively. 
Furthermore, EventDC incorporates two key components: Event-Modulated Alignment (EMA) and Local Depth Filtering (LDF). 
At each encoder stage, EMA predicts spatial offsets from event features and uses them to adjust the pixel distributions of RGB and depth features. This enables more precise multi-modal alignment and fusion. In addition, a structure-aware loss is introduced to further enhance the consistency. 
At the decoder stage, LDF leverages event features to estimate motion masks that identify moving objects. It then refines the depth values within these regions using dynamic convolutions and a local motion-aware loss, ultimately enhancing depth accuracy around the moving objects.

\textbf{Event-Modulated Alignment.} 
As depicted in Fig.~\ref{fig_pipeline}, at the $j$-th ($j\in \{ 1,2,3,4 \}$) stage of the three encoders, the EMA module takes as input the color image feature $\mathbf{I}_j$, sparse depth feature $\mathbf{S}_j$, and event feature $\mathbf{E}_j$, each with dimensions $\mathbb{R}^{C \times H \times W}$, where $C$, $H$, and $W$ denote the channel, height, and width, respectively. These inputs are first individually processed by three separate $3 \times 3$ convolutional layers $\mathcal{F}_{{\tau}_{j1}}(\cdot)$, $\mathcal{F}_{{\tau}_{j2}}(\cdot)$ and $\mathcal{F}_{{\tau}_{j3}}(\cdot)$, with a stride of 1 and output channels of $C$, $2C$, and $C$, respectively. The transformed event feature is then fused with the transformed RGB and depth features, respectively, resulting in the intermediate features:
\begin{subequations}
\begin{align}
    &\bar{\mathbf{Q}}_{j}=\mathcal{F}_{{\tau}_{j1}} \left( \mathbf{I}_j \right ) + \alpha \cdot \mathcal{F}_{s} \left ( \mathcal{F}_{{\tau}_{j2}} \left( \mathbf{E}_j \right ) \right), \\
    &\tilde{\mathbf{Q}}_{j}=\mathcal{F}_{{\tau}_{j3}} \left( \mathbf{S}_j \right ) + \beta \cdot \mathcal{F}_{s} \left ( \mathcal{F}_{{\tau}_{j2}} \left( \mathbf{E}_j \right ) \right ),
\end{align}\label{eq_pre_offset}
\end{subequations}
where $\mathcal{F}_{s}(\cdot)$ denotes the operation that splits the $2C$-channel feature into two $C$-channel parts. $\alpha$ and $\beta$ are learnable terms \footnote{Implemented using \texttt{torch.nn.Parameter(zeros(1))} with the zero initialization designed to facilitate the progressive learning of event priors during training.} that control the contribution of the event term. 

Subsequently, these two intermediate features are used to predict the offsets via two additional $3\times 3$ convolutions, $\mathcal{F}_{{\tau}_{j4}}(\cdot)$ and $\mathcal{F}_{{\tau}_{j5}}(\cdot)$, producing $2K\times H\times W$ offsets and $K\times H\times W$ weights:
\begin{subequations}
\begin{align}
    &\bar{\mathbf{o}}_{j}, \bar{\mathbf{w}}_{j}=\mathcal{F}_{{\tau}_{j4}} ( \bar{\mathbf{Q}}_{j} ), \\
    &\tilde{\mathbf{o}}_{j}, \tilde{\mathbf{w}}_{j}=\mathcal{F}_{{\tau}_{j5}} ( \tilde{\mathbf{Q}}_{j} ).
\end{align}\label{eq_offset}
\end{subequations}
This step enables the model to adaptively determine the sampling locations by learning a prior from event data that is sensitive to fast motion. Consequently, the dynamic convolution in Eq.~\eqref{eq_dcn} can be used to perform pixel-wise adjustment of image and depth features formulated as follows:
\begin{subequations}
\begin{align}
&\hat{\mathbf{I}}_j = \mathcal{F}_{\psi}(\mathbf{I}_j; \ \bar{\mathbf{o}}_{j}, \bar{\mathbf{w}}_{j}),\\
&\hat{\mathbf{S}}_j = \mathcal{F}_{\psi}(\mathbf{S}_j; \ \tilde{\mathbf{o}}_{j}, \tilde{\mathbf{w}}_{j}),
\end{align}\label{eq_realign}
\end{subequations}
where $\mathcal{F}_{\psi}(\cdot)$ denotes the generalized form of the operation defined in Eq.~\eqref{eq_dcn}. Note that Eq.\eqref{eq_realign} emphasizes RGB-D pixel redistribution under highly dynamic conditions with offsets instead of adaptive weights. Consequently, in contrast to Eq.\eqref{eq_dcn}, both $\bar{\mathbf{w}}_{j}$ and $\tilde{\mathbf{w}}_{j}$ correspond to the predefined weights $\mathbf{w}$ with $\Delta \mathbf{m}$ being the identity matrix. Subsequently, the redistributed RGB-D features are further processed by a $3 \times 3$ convolution $\mathcal{F}_{{\tau}_{j6}}(\cdot)$ to obtain the fused feature $\mathbf{F}_j \in \mathbb{R}^{C \times H \times W}$, which is formulated as:
\begin{equation}
\mathbf{F}_j = \mathcal{F}_{{\tau}_{j6}} ( \hat{\mathbf{I}}_j + \hat{\mathbf{S}}_j ).
\label{eq_align_fuse}
\end{equation}
Additionally, a structure-aware loss $\mathcal{L}_{str}$ is introduced to enhance the consistency. Let $\mathcal{G}(\cdot)$ denote a sequence of single-channel convolution, Min-Max normalization, and gradient computation:
\begin{equation}
\mathcal{L}_{str} = \sum_{j=1}^{4} \frac{1}{n} \| \mathcal{G}(\hat{\mathbf{I}}_j) - \mathcal{G}(\hat{\mathbf{S}}_j) \|_2^2.
\label{eq_loss_str}
\end{equation}

\textbf{Local Depth Filtering.} 
As shown in Fig.\ref{fig_pipeline}, the LDF module takes the depth feature $\mathbf{D}_i$ and event feature $\mathbf{E}_i$ as inputs to adaptively generate offsets and weights at the $i$-th stage of the decoder ($i \in \{1,2,3\}$). Following the strategy used in Eqs.\eqref{eq_pre_offset}–\eqref{eq_realign}, this results in the updated depth feature: 
\begin{equation}
\hat{\mathbf{D}}_i = \mathcal{F}_{\psi}(\mathbf{D}_i; \ \tilde{\mathbf{o}}_{i}, \tilde{\mathbf{w}}_{i}).
\label{eq_ldf_dcn}
\end{equation}
In contrast, the modulation scalar $\Delta m$ within $\tilde{\mathbf{w}}_{i}$ is learned jointly from the depth and event inputs. 
Furthermore, to explicitly model regions of dynamic objects, LDF predicts a motion mask $m_i$ based on $\mathbf{E}_i$ using a sigmoid activation $\sigma(\cdot)$ after a single-channel $3 \times 3$ convolution $\mathcal{F}_{{\tau}_{i6}}(\cdot)$:
\begin{equation}
m_i = \mathcal{\sigma}(\mathcal{F}_{{\tau}_{i6}}\left ( \mathbf{E}_i \right )).
\label{eq_ldf_mask}
\end{equation}
By combining Eqs.~\eqref{eq_ldf_dcn} and~\eqref{eq_ldf_mask}, LDF refines depth with a focus on dynamic regions to get:
\begin{equation}
\mathring{\mathbf{D}}_i = m \cdot \hat{\mathbf{D}}_i + (1-m) \cdot \mathbf{D}_i.
\label{eq_ldf_update}
\end{equation}

Finally, the output $\mathring{\mathbf{D}}_1$ from the last LDF module is passed through a $3 \times 3$ convolutional tail $\mathcal{F}_{{\tau}_t}(\cdot)$ to generate the dense depth prediction: 
\begin{equation}
\mathbf{D} = \mathcal{F}_{{\tau}_t}(\mathring{\mathbf{D}}_1).
\label{eq_prediction}
\end{equation}

Additionally, we introduce a motion-aware loss to enhance the depth recovery around motion areas:
\begin{equation}
\mathcal{L}_{mot} = \sum_{i=1}^{3} \frac{1}{n} \| b_i \cdot \mathcal{H}(\mathring{\mathbf{D}}_i) - b_i \cdot \mathcal{F}_{d}(\mathbf{Z}) \|_2^2,
\label{eq_loss_mot}
\end{equation}
where $\mathbf{Z}$ is the GT depth, $\mathcal{H}(\cdot)$ applies ReLU and a single-channel convolution, and $b_i$ is a binary mask with $b_i = 1$ if $m_i$ exceeds its mean, and $0$ otherwise. $\mathcal{F}_d(\cdot)$ denotes the downsampling operation.

\textbf{Discussion.} 
In summary, EMA and LDF differ from previous dynamic convolution methods in two key aspects:
\textbf{(1)} 
Unlike traditional methods that typically rely on single-modal and single-path inputs, our approach adopts a multi-modal and multi-path input design, where key convolutional parameters are derived from different modalities.
\textbf{(2)} 
Our method is data-driven where we use event-based adaptation to address global misalignment and local depth inaccuracies caused by fast motion.

\vspace{-5pt}
\subsection{Loss Function}
Given the predicted depth $\mathbf{D}$ and GT depth $\mathbf{Z}$ with $n$ valid pixels, we adopt a commonly used reconstruction loss \cite{park2020nonlocal,lin2022dynamic,zhang2023cf,wang2023lrru,wang2025learning} to formulate the training objective: 
\begin{equation}
    \mathcal{L}_{rec} = \frac{1}{n} ( \left \| \mathbf{D} - \mathbf{Z} \|_2^2 + \| \mathbf{D} - \mathbf{Z} \|_1 \right )
    \label{eq_loss_rec}.
\end{equation}

By combining the reconstruction loss with the structure-constrained loss $\mathcal{L}_{str}$ in Eq.~\ref{eq_loss_str} and motion-aware loss $\mathcal{L}_{rec}$ in Eq.~\ref{eq_loss_mot}, the overall loss function is formulated as:
\begin{equation}
\mathcal{L}_t = \mathcal{L}_{rec} + \lambda \mathcal{L}_{str} + \mu \mathcal{L}_{mot},
\label{eq_loss_total}
\end{equation}
where $\lambda$ and $\mu$ are weighting hyper-parameters that we empirically set to 1 and 0.1, respectively.

\begin{table}[t]
\caption{
Basic statistics of the EventDC benchmark.
}\label{tab_dataset_info}
\vspace{0pt}
\centering
\Large
\renewcommand\arraystretch{1.12}
\resizebox{0.962\textwidth}{!}{
\begin{tabular}{l|c|c|c|cc|c}
\toprule[1.8pt]
Dataset          & Color Camera       & Depth Sensor            & Event Camera  & Train   & Test   & Resolution         \\ 
\midrule
EventDC-Real     & FILR BFS-U3-31S4C  & Ouster OS1-128 LiDAR    &  DAVIS346     & 14,845  & 1,000  & $320 \times 256$   \\
EventDC-SemiSyn  & PointGrey Flea2    & Velodyne HDL-64E LiDAR  &  -            & 7,094   & 2,213  & $1216 \times 256$  \\
EventDC-FullSyn  & -                  & -                       & -             & 21,000  & 500    & $512 \times 256$   \\
\bottomrule[1.8pt]
\end{tabular}
}
\vspace{-10pt}
\end{table}

\begin{figure}[t]
\centering
\includegraphics[width=0.95\columnwidth]{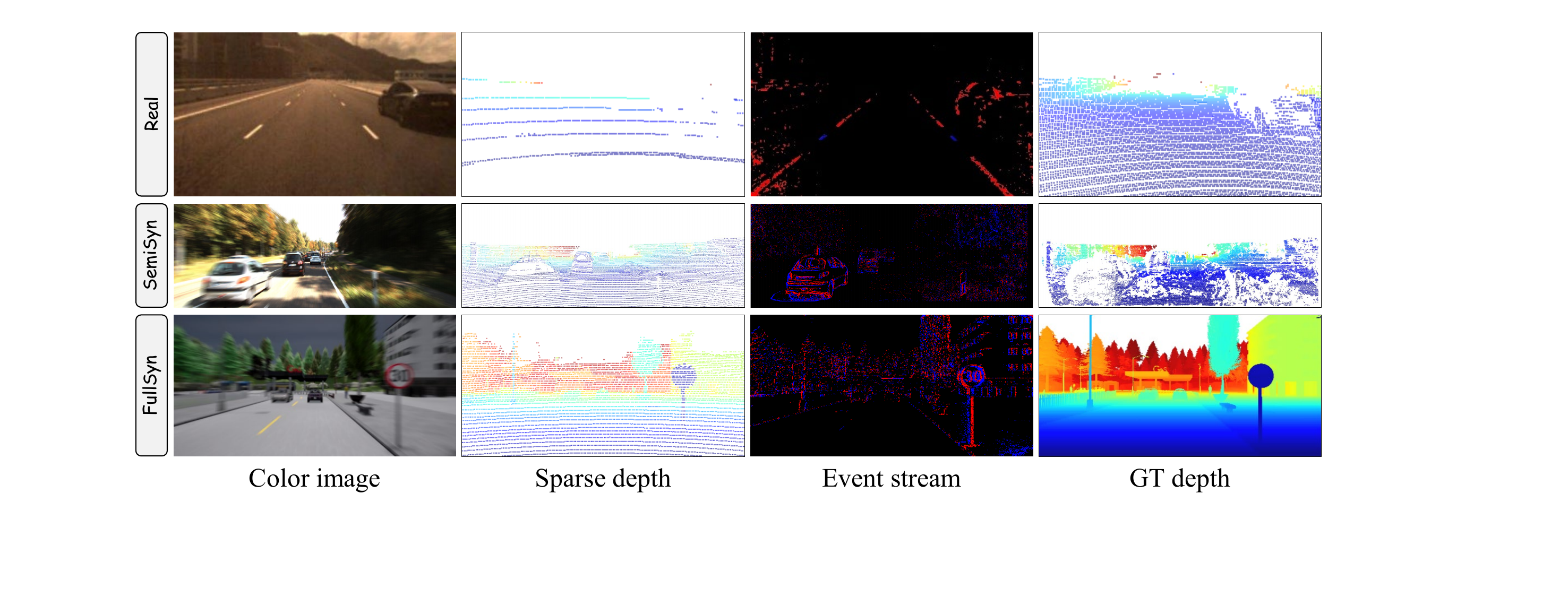}\\
\vspace{-5pt}
\caption{Visualizations of the proposed EventDC benchmark: EventDC-Real/SemiSyn/FullSyn.
}
\label{fig_data_case}
\vspace{-12pt}
\end{figure}

\section{EventDC Benchmark}
\textbf{Motivation.} 
Traditional depth completion datasets \cite{geiger2012we,silberman2012indoor,yan2024tri,song2015sun} rely on the fusion of color images and sparse depth maps to predict dense depth. However, this approach suffers in highly dynamic environments especially when dealing with fast ego-motion and object motion. 
This is due to unreliable low-frame-rate RGB images and sparse depth data from motion blur and sampling inconsistencies. 
Event cameras with the capability to capture high temporal resolution and sensitivity to rapid movements \cite{gallego2020event} provide an ideal solution to overcome these limitations. By asynchronously recording minute brightness variations, event cameras can offer accurate depth information in dynamic scenarios where conventional RGB-D sensors fail. In light of these characteristics, we propose an event-based depth completion benchmark that leverages the unique advantages of event data to address the challenges of depth completion in dynamic environments.

\textbf{Data Collection.} 
Tab.~\ref{tab_dataset_info} provides an overview of the sensors used in the datasets with their respective specifications. 
\textit{EventDC-Real} is extended from \cite{wei2024fusionportablev2}, a real-world dataset in which color images and event frames are captured using the FLIR BFS-U3-31S4C camera and the DAVIS346 sensor, respectively. The GT depth is acquired from a 128-line Ouster LiDAR, and the sparse depth is derived from its 16 sub-lines. 
\textit{EventDC-SemiSyn} is a semi-synthetic dataset based on KITTI \cite{geiger2012we}. The sparse depth and GT depth come from the raw data of KITTI. For the color images, we apply radial motion blur by progressively scaling and transforming the image around its center to simulate a motion blur effect with adjustable strength and step count. Additionally, VID2E \cite{gehrig2020video} is used to generate the event data with frames captured within 15 ms before and after the current timestamp. 
\textit{EventDC-FullSyn} is a fully synthetic dataset generated using the CARLA simulator \cite{dosovitskiy2017carla}. The color images are processed similarly with radial motion blur. 
Finally, to facilitate model training, the resolution of all datasets has been cropped to multiples of 32. Fig.~\ref{fig_data_case} presents visual examples from these three datasets.

\section{Experiment}
\textbf{Metric and Implementation Detail.} 
Following previous depth completion methods \cite{geiger2012we, tang2020learning, yan2024tri, liang2025distilling}, we adopt RMSE (mm), MAE (mm), REL, and threshold accuracy $\delta$ (\%) as evaluation metrics. Refer to the appendix for their full definitions. 
We implement EventDC using the PyTorch framework and conduct training on two NVIDIA RTX 4090 GPUs using the Distributed Data Parallel strategy for efficiency. Optimization is performed with the AdamW optimizer \cite{loshchilov2017decoupled} in conjunction with the OneCycle learning rate policy \cite{smith2019super}. The training process begins with a warm-up stage that linearly increases the learning rate from 0.00002 to 0.001 over the first 10\% of iterations. Subsequently, a cosine annealing schedule gradually decays the learning rate to a final value of 0.0002. The batch size is set to 2 per GPU. In addition, to further enhance model performance, we employ a set of data augmentation strategies \cite{tang2020learning,liu2021fcfr}, including random horizontal flip, rotation, cropping, and color jitter.

\begin{table}[t]
\small
\caption{Quantitative depth completion comparisons on the EventDC-Real dataset.}
\label{tab_real}
\vspace{2pt}
\centering
\renewcommand\arraystretch{1.12}
\resizebox{0.89\textwidth}{!}{
\begin{tabular}{l|cccccc|c}
\toprule[1.8pt]
{Method}                   & RMSE  $\downarrow$     & MAE  $\downarrow$  & REL $\downarrow$   & $\delta _{1.05}$  $\uparrow$  & $\delta _{1.10}$  $\uparrow$  & $\delta _{1.15}$  $\uparrow$  & Venue \\ 
\midrule
CSPN \cite{2018Learning}          &858.5 &284.6 &0.0386 &90.0 &94.4 &96.0 &ECCV 2018\\
S2D \cite{ma2018sparse}           &984.1 &410.8 &0.0565 &82.1 &90.3 &93.4 &ICRA 2018\\
FusionNet \cite{vangansbeke2019}  &\underline{658.1} &262.4 &0.0384 &87.6 &93.9 &96.0 &MVA 2019\\
RigNet \cite{yan2022rignet} &685.4 &234.6 &0.0336 &87.5 &92.3 &95.4 &ECCV 2022\\
DySPN \cite{lin2022dynamic}       &700.1 &223.7 &0.0285 &91.3 &95.1 &96.6 &AAAI 2022\\
Prompting \cite{park2024depth}    &670.7 &\underline{205.1} &\underline{0.0252} &\underline{92.1} &\underline{95.7} &\underline{97.0} &CVPR 2024\\
OGNI-DC \cite{zuo2024ogni}        &709.7 &231.1 &0.0294 &90.9 &94.8 &96.4 &ECCV 2024 \\
SigNet \cite{yan2024completion}   & 906.4 & 348.1 &0.0345 & 83.5 & 90.6 & 93.2 & CVPR 2025 \\
LPNet \cite{wang2025learning}     &911.2 &389.0 &0.0472 &83.6 &90.4 &93.3 & arXiv 2025\\
\midrule
\textbf{EventDC (our)}            &\textbf{574.0} &\textbf{179.0} &\textbf{0.0242} &\textbf{92.9} &\textbf{96.3} &\textbf{97.5} &-\\
\textit{Improvement} \ $\uparrow$  & \textcolor{teal}{\textit{84.1}}  &\textcolor{teal}{\textit{26.1}}  &\textcolor{teal}{\textit{0.0010}}  &\textcolor{teal}{\textit{0.8}}  &\textcolor{teal}{\textit{0.6}}  &\textcolor{teal}{\textit{0.5}} & -\\
\bottomrule[1.8pt]
\end{tabular}
}
\vspace{-6pt}
\end{table}

\begin{figure}[t]
\centering
\includegraphics[width=0.882\columnwidth]{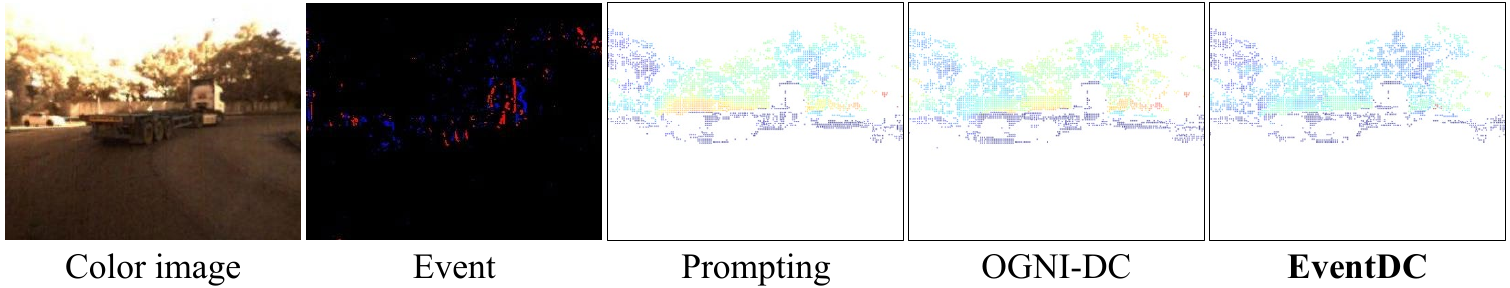}\\
\vspace{-2pt}
\caption{Depth error comparisons on EventDC-Real. Warmer color indicates higher error.
}
\label{fig_vis_real}
\vspace{-5pt}
\end{figure}

\subsection{Comparisons with State-of-the-arts}
In this section, we compare our EventDC with well-known methods: 
CSPN \cite{2018Learning}, S2D \cite{ma2018sparse}, FusionNet \cite{vangansbeke2019}, RigNet \cite{yan2022rignet}, DySPN \cite{lin2022dynamic}, Prompting \cite{park2024depth}, OGNI-DC \cite{zuo2024ogni}, SigNet \cite{yan2024completion}, and LPNet \cite{wang2025learning}. For a fair comparison, we retrain all methods from scratch on the proposed benchmark. Note that BPNet \cite{tang2024bilateral}, TPVD \cite{yan2024tri}, and DMD$^3$C \cite{liang2025distilling} are excluded from the comparison. This is because they require additional camera parameters during training which are not available in our settings.

\textbf{EventDC-Real.} 
We first evaluate the proposed EventDC on EventDC-Real, a real-world dataset collected using various devices such as handheld sensors and robotic platforms. 
The numerical results are summarized in Tab~\ref{tab_real}. 
Our EventDC achieves the overall lowest errors while maintaining the highest accuracy across the board. For example, it outperforms the second-best method by 84.1 mm in RMSE, 26.1 mm in MAE, 0.001 in REL, and 0.8 points in $\delta_{1.05}$. 
Compared to post-refinement methods such as CSPN\cite{2018Learning} and DySPN\cite{lin2022dynamic}, our EventDC without any post-processing consistently achieves better performance. 
Even when compared to the large-scale depth foundation model Promoting\cite{park2024depth}, our approach achieves superior results with significantly fewer model parameters. 
Fig.~\ref{fig_vis_real} presents the comparisons of depth error. It clearly shows that our EventDC produces more accurate depth results especially around moving objects.

\textbf{EventDC-SemiSyn.} 
To further validate the effectiveness of EventDC, we evaluate it on EventDC-SemiSyn, a semi-synthetic dataset comprising synthetically generated event frames and color images rendered under highly dynamic conditions. 
As reported in Tab.~\ref{tab_semi}, our EventDC continues to deliver outstanding results across all evaluation metrics. 
On average, it outperforms recent methods: 
Prompting~\cite{park2024depth}, OGNI-DC~\cite{zuo2024ogni}, and LPNet~\cite{wang2025learning} by 18.9\%, 30.8\%, and 27.7\% in RMSE, MAE, and REL, respectively, and by 3.9, 1.6, and 0.9 percentage points in $\delta_{1.05}$, $\delta_{1.10}$, and $\delta_{1.15}$, respectively. 
As illustrated in Fig.~\ref{fig_vis_semi}, our EventDC effectively reconstructs accurate depth details and structural consistency even under highly dynamic scenes.

\textbf{EventDC-FullSyn.} 
Apart from the real and semi-synthetic settings, we also validate EventDC on the fully synthetic dataset, EventDC-FullSyn, to further assess its generalization capability under diverse scenarios. As shown in Tab.\ref{tab_full}, our EventDC consistently outperforms all competing approaches by large margins. For example, it surpasses the second-best approach by 53.5 mm in RMSE and 19.4 mm in MAE. In addition, it achieves a 13.0\% improvement in REL compared to the foundation model-based Prompting \cite{park2024depth}. These results demonstrate the robustness of our EventDC in reducing both absolute and relative errors. 
Fig.~\ref{fig_vis_full} shows that our EventDC yields more refined details and sharper object boundaries than others, which highlight its effectiveness in fully synthetic scenarios.

\begin{table}[t]
\centering
\begin{minipage}{0.72\linewidth}
    \caption{Quantitative comparisons on the EventDC-SemiSyn dataset.}
    \label{tab_semi}
    \vspace{2pt}
    \renewcommand\arraystretch{1.12}
    \resizebox{1\textwidth}{!}{
    \begin{tabular}{l|cccccc}
    \toprule[1.8pt]
    {Method} & RMSE$\downarrow$ & MAE$\downarrow$ & REL$\downarrow$ & $\delta_{1.05}\uparrow$ & $\delta_{1.10}\uparrow$ & $\delta_{1.15}\uparrow$ \\
    \midrule
    CSPN \cite{2018Learning} & 989.8 & 262.8 & 0.0189 & 94.6 & 97.2 & 98.1 \\
    S2D \cite{ma2018sparse} & 1097.3 & 366.4 & 0.0237 & 91.0 & 96.4 & 97.9 \\
    FusionNet \cite{vangansbeke2019} & 877.6 & 333.1 & 0.0258 & 92.6 & \underline{98.2} & \underline{98.8} \\
    RigNet \cite{yan2022rignet} & 858.2 & 216.4 & 0.0156 & 95.1 & 97.8 & 98.1 \\
    DySPN \cite{lin2022dynamic} &897.7 &\underline{207.5} &0.0149 &\underline{95.9} &97.8 &98.6 \\
    Prompting \cite{park2024depth} & 873.9 & 291.1 & 0.0198 & 92.6 & 97.1 & 98.4 \\
    OGNI-DC \cite{zuo2024ogni} & \underline{832.0} & 210.5 & \underline{0.0143} & 95.7 & 98.0 & 98.7 \\
    SigNet \cite{yan2024completion} & 1065.4 & 321.3 & 0.0226 & 91.1 & 97.0 & 98.1 \\
    LPNet \cite{wang2025learning} & 1283.4 & 416.3 & 0.0242 & 90.0 & 95.3 & 97.2 \\
    \midrule
    \textbf{EventDC (ours)} & \textbf{778.8} & \textbf{196.2} & \textbf{0.0134} & \textbf{96.7} & \textbf{98.4} & \textbf{99.0} \\
    \textit{Improvement} $\uparrow$ & \textcolor{teal}{\textit{53.2}} & \textcolor{teal}{\textit{11.3}} & \textcolor{teal}{\textit{0.0009}} & \textcolor{teal}{\textit{0.8}} & \textcolor{teal}{\textit{0.2}} & \textcolor{teal}{\textit{0.2}} \\
    \bottomrule[1.8pt]
    \end{tabular}
    }
\end{minipage}\hfill
\begin{minipage}{0.256\linewidth}
\vspace{15pt}
    \centering
    \includegraphics[width=\linewidth]{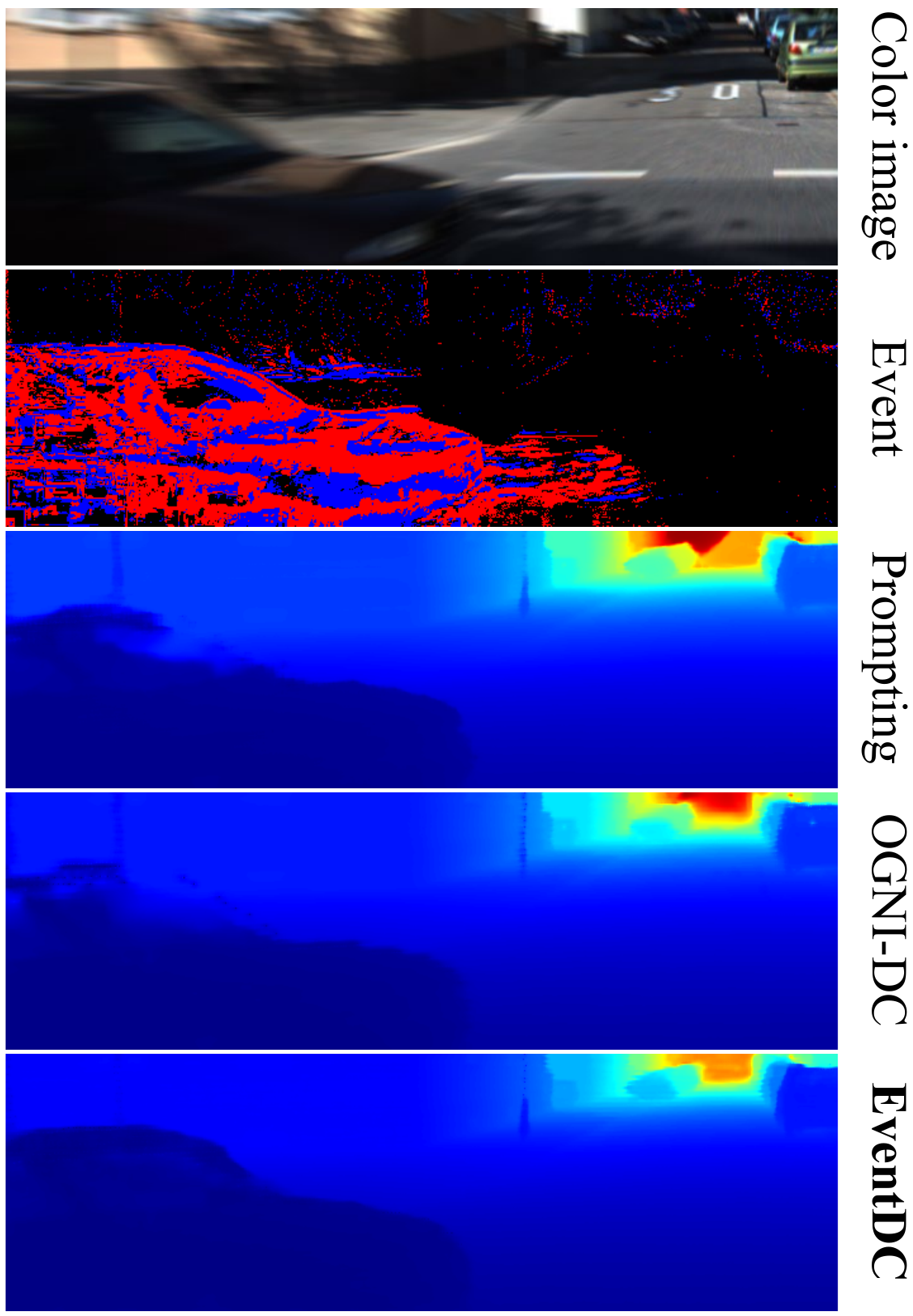}
    \vspace{-10pt}
    \captionof{figure}{Visual results.}
    \label{fig_vis_semi}
\end{minipage}
\vspace{-26pt}
\end{table}

\begin{table}[t]
\centering
\begin{minipage}{0.72\linewidth}
    \caption{Quantitative comparisons on the EventDC-FullSyn dataset.}
    \label{tab_full}
    \vspace{2pt}
    \renewcommand\arraystretch{1.12}
    \resizebox{1\textwidth}{!}{
    \begin{tabular}{l|cccccc}
    \toprule[1.8pt]
    {Method} & RMSE$\downarrow$ & MAE$\downarrow$ & REL$\downarrow$ & $\delta_{1.05}\uparrow$ & $\delta_{1.10}\uparrow$ & $\delta_{1.15}\uparrow$ \\
    \midrule
    CSPN \cite{2018Learning} & 864.9 & 399.5 & 0.1193 & 62.7 & 80.9 & 87.6 \\
    S2D \cite{ma2018sparse} & 899.0 & 376.2 & 0.1243 & 69.8 & 83.8 & 89.1 \\
    FusionNet \cite{vangansbeke2019} & 670.6 & 230.9 & 0.0931 & 77.3 & 86.6 & 90.4 \\
    RigNet \cite{yan2022rignet} & 723.4 & 166.3 & 0.0578 & 81.1 & 91.6 & 92.8 \\
    DySPN \cite{lin2022dynamic} & 679.8 & 165.6 & 0.0646 & 87.2 & 92.6 & 94.6 \\
    Prompting \cite{park2024depth} & 709.7 & 180.9 & \underline{0.0538} & \underline{90.7} & \underline{93.8} & \underline{95.3} \\
    OGNI-DC \cite{zuo2024ogni} & \underline{673.7} & \underline{162.5} & 0.0578 & 87.8 & 93.0 & 95.1 \\
    SigNet \cite{yan2024completion} & 904.5 & 349.2 & 0.0902 & 76.3 & 84.1 & 90.3 \\
    LPNet \cite{wang2025learning} & 920.2 & 357.3 & 0.0943 & 75.1 & 85.9 & 90.3 \\
    \midrule
    \textbf{EventDC (ours)} & \textbf{620.2} & \textbf{143.1} & \textbf{0.0468} & \textbf{92.1} & \textbf{95.5} & \textbf{96.8} \\
    \textit{Improvement} $\uparrow$ & \textcolor{teal}{\textit{53.5}} & \textcolor{teal}{\textit{19.4}} & \textcolor{teal}{\textit{0.0070}} & \textcolor{teal}{\textit{1.4}} & \textcolor{teal}{\textit{1.7}} & \textcolor{teal}{\textit{1.5}} \\
    \bottomrule[1.8pt]
    \end{tabular}
    }
\end{minipage}\hfill
\begin{minipage}{0.256\linewidth}
\vspace{15pt}
    \centering
    \includegraphics[width=\linewidth]{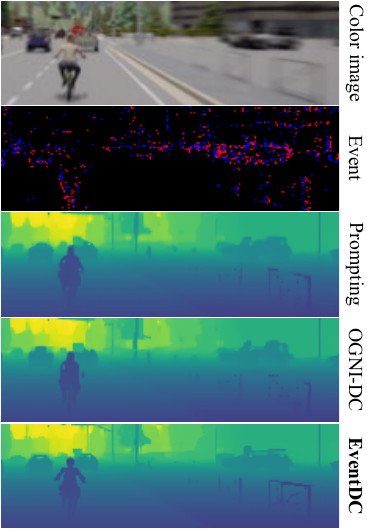}
    \vspace{-10pt}
    \captionof{figure}{Visual results.}
    \label{fig_vis_full}
\end{minipage}
\vspace{-18pt}
\end{table}

\begin{wraptable}{r}{0.474\textwidth}
\Large
\caption{Complexity on EventDC-Real.}
\label{tab_complexity}
\vspace{-5pt}
\centering
\renewcommand\arraystretch{1.12}
\resizebox{0.47\textwidth}{!}{
\begin{tabular}{l|ccc|c}
\toprule[2.5pt]
\multirow{2}{*}{Method}        & Param.  & Memo.  & Time  & RMSE \\
                               & (M) $\downarrow$ & (GB) $\downarrow$  & (ms) $\downarrow$  & (mm) $\downarrow$  \\
\midrule
DySPN \cite{lin2022dynamic}    & \textbf{26.3}  & \textbf{0.9}  & \textbf{9.8}  & 700.1  \\
RigNet \cite{yan2022rignet}    & 65.2  & 2.3  & 26.5  & 685.4  \\
Prompting \cite{park2024depth} & 326.9 & 4.1  & 39.5   & \underline{670.7} \\
OGNI-DC \cite{zuo2024ogni}     & 84.4  & 3.7  & 314.1  & 709.7  \\
LPNet \cite{wang2025learning}  & \underline{29.6}  & \underline{1.1} & \underline{18.4}  & 911.2 \\
\midrule
\textbf{EventDC (our)}         & 43.2  &1.5  & 41.5 & \textbf{574.0} \\
\bottomrule[2.5pt]
\end{tabular}
}
\end{wraptable}

\textbf{Complexity Analysis.} 
Tab.~\ref{tab_complexity} presents the complexity comparisons between our EventDC and other competing methods in terms of model parameters (Param.), memory consumption (Memo.), and inference time. Our EventDC not only achieves outstanding performance, but also maintains competitive efficiency. In particular, compared to the second-best method Prompting \cite{park2024depth}, our EventDC achieves a significantly lower RMSE by 96.7 mm with only about one-eighth the number of parameters and one-third the memory.

\subsection{Ablation Studies}
Tab.~\ref{tab_ablation_eventdc} summarizes the ablation results on EventDC-Real. EventDC-i serves as a UNet-style baseline that takes only sparse depth as input and employs additive skip connections.

\textbf{(1)} 
EventDC-ii further develops this approach by utilizing RGB images and integrating RGB-D features through additive fusion. Although depth input is sparse, RGB offers rich structural and semantic details. 
This leads to a significant decrease in error and substantial gains in accuracy. For example, the RMSE is reduced by 41.3 mm and the MAE by 32.1 mm. 
EventDC-iii enhances support for event streams, which are advantageous because of their fine temporal detail and motion sensitivity. Consequently, this makes them very effective in dynamic environments where they supplement depth data. 
EventDC-iv combines all three modalities to give consistent improvements across all evaluation metrics. Specifically, it surpasses the baseline by 12.2\%, 20.7\%, and 8.5\% in RMSE, MAE, and REL, respectively. It concurrently improves $\delta_{1.05}$, $\delta_{1.10}$, and $\delta_{1.15}$ by 1.0, 0.3, and 0.2 percentage points. 
These results underscore the effectiveness of multi-modal fusion, 
where the integration of complementary modalities enables more accurate and complete depth reconstruction.

\begin{figure}[t]
\centering
\includegraphics[width=0.975\columnwidth]{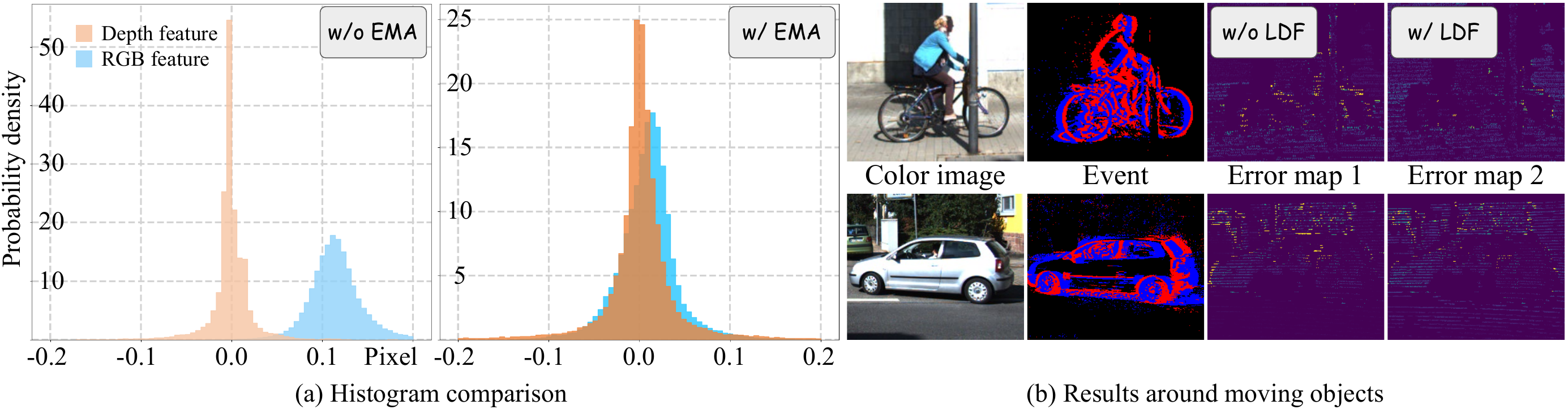}\\
\vspace{-5pt}
\caption{
Statistical and visual comparative analyses of the proposed EMA and LDF modules. 
}
\label{fig_vis_ablation}
\vspace{-11pt}
\end{figure}

\textbf{(2)} 
EventDC-v to EventDC-ix conduct ablation studies to examine the impact of dynamic convolution (DConv), EMA, and LDF in the encoder (Enc) and decoder (Dec) stages. 
Specifically, the introduction of DConv in EventDC-v brings notable benefits. Furthermore, EventDC-vi with EMA further reduces RMSE by 26 mm. These results validate the efficacy of our event-based adaptive alignment strategy. 
Fig.~\ref{fig_vis_ablation}(a) compares the distributions of RGB-D features with and without EMA. 
EMA works as intended in promoting better alignment between the two modalities with more consistent feature representations. 
Similarly, EventDC-viii with LDF outperforms EventDC-vii with DConv by 25.2 mm. This demonstrates its superior ability to recover fine-grained local depth which is further evident in Fig.~\ref{fig_vis_ablation}(b). 
Finally, EventDC-ix which integrates both EMA and LDF modules achieves the best overall performance. It reduces RMSE by 16.0\% (from 683.3 mm) and MAE by 20.3\% (from 224.7 mm). In summary, each component contributes positively to the overall performance gains.

\begin{table}[t]
\large
\caption{Ablations on EventDC-Real. DConv/Enc/Dec: dynamic convolution/encoder/decoder.}
\label{tab_ablation_eventdc}
\vspace{0pt}
\centering
\renewcommand\arraystretch{1.1}
\resizebox{0.98\textwidth}{!}{
\begin{tabular}{l|ccc|cc|cc|cccccc}
\toprule[1.8pt]
\multirow{2}{*}{EventDC} & \multicolumn{3}{c|}{Modality} & \multicolumn{2}{c|}{DConv} & EMA & LDF & \multirow{2}{*}{RMSE} & \multirow{2}{*}{MAE} & \multirow{2}{*}{REL}  & \multirow{2}{*}{$\delta _{1.05}$}  & \multirow{2}{*}{$\delta _{1.10}$}  & \multirow{2}{*}{$\delta _{1.15}$}  \\ 
& Depth  & RGB  & Event  & Enc & Dec & Enc & Dec  &   &  & & & &    \\ 
\midrule
i   & \checkmark  &             &      &       &      &       &             &727.2 &283.2 &0.0341 &89.3 &94.4 &96.2  \\
ii  & \checkmark  & \checkmark  &     &        &      &       &             &685.9 &251.1 &0.0328 &89.8 &94.5 &96.2  \\
iii & \checkmark  &  & \checkmark             &    &  &       &             &696.0 &244.3 &0.0314 &90.0 &94.5 &96.3  \\
iv & \checkmark  & \checkmark  & \checkmark  &    &  &       &              &638.3 &224.7 &0.0312 &90.3 &94.7 &96.4  \\
\midrule
v  & \checkmark  & \checkmark  & \checkmark  & \checkmark & & &   & 628.8 & 219.5 & 0.0292 & 91.0 &95.0 &96.7  \\
vi  & \checkmark  & \checkmark  & \checkmark  & & & \checkmark  &             &602.8 &196.1 &0.0276 & 91.7 &95.6 &97.1  \\
vii  & \checkmark  & \checkmark  & \checkmark  & & \checkmark & &   & 630.6 & 219.8 & 0.0295 & 91.0 & 94.8 & 96.5  \\
viii  & \checkmark  & \checkmark  & \checkmark  & & & & \checkmark   & 605.4 & 198.3 &0.0279 & 91.5 & 95.6 & 97.1 \\
ix   & \checkmark  & \checkmark  & \checkmark  & & & \checkmark  & \checkmark  &574.0 &179.0 &0.0242 &92.9 &96.3 &97.5  \\ 
\bottomrule[1.8pt]
\end{tabular}
}
\vspace{-9pt}
\end{table}

\section{Conclusion} \vspace{-1mm}
We propose EventDC in this work. Our EventDC is the first depth completion framework that tackles the challenges of dynamic scenes by harnessing the unique strengths of event data. 
To mitigate the adverse effects of fast ego-motion and object motion, our EventDC incorporates two event-driven modules: event-modulated alignment and local depth filtering. These modules, supported by two dedicated loss constraints, address global misalignment and local depth inaccuracies, respectively. 
To further support research in this area, we construct the first benchmark for event-based depth completion comprising one real-world and two synthetic datasets. 
Extensive experiments demonstrate the effectiveness of our EventDC and its superior performance in challenging dynamic environments.

\textbf{Limitation and Broader Impact.} 
Despite achieving promising results in dynamic scenes, our EventDC relies on high-quality event data and precise sensor alignment that may not be easily attainable in all real-world settings. 
The EMA and LDF modules introduce additional computational costs, potentially limiting deployment on resource-constrained devices. Moreover, the scale and diversity of our real-world dataset are limited, and future work is needed to evaluate generalization across more diverse environments and motion patterns. 
Despite these limitations, our EventDC offers a step forward in robust depth perception under motion blur and rapid dynamics with potential applications in autonomous driving, robotics, AR/VR, \textit{etc}. By introducing a dedicated benchmark, we aim to promote research in event-based depth completion. As with all perceptual systems, responsible deployment requires attention to reliability, fairness, and safety in complex real-world conditions.

{   
\bibliographystyle{ieeenat_fullname}
\bibliography{ref}

\begin{thebibliography}{73}
\providecommand{\natexlab}[1]{#1}
\providecommand{\url}[1]{\texttt{#1}}
\expandafter\ifx\csname urlstyle\endcsname\relax
  \providecommand{\doi}[1]{doi: #1}\else
  \providecommand{\doi}{doi: \begingroup \urlstyle{rm}\Url}\fi

\bibitem[Chen et~al.(2019)Chen, Yang, Liang, and Urtasun]{chen2019learning}
Yun Chen, Bin Yang, Ming Liang, and Raquel Urtasun.
\newblock Learning joint 2d-3d representations for depth completion.
\newblock In \emph{ICCV}, pages 10023--10032, 2019.

\bibitem[Cheng et~al.(2018)Cheng, Wang, and Yang]{2018Learning}
Xinjing Cheng, Peng Wang, and Ruigang Yang.
\newblock Learning depth with convolutional spatial propagation network.
\newblock In \emph{ECCV}, pages 103--119, 2018.

\bibitem[Cheng et~al.(2019)Cheng, Wang, and Yang]{cspn_pami}
Xinjing Cheng, Peng Wang, and Ruigang Yang.
\newblock Learning depth with convolutional spatial propagation network.
\newblock \emph{IEEE Transactions on Pattern Analysis and Machine Intelligence}, 42\penalty0 (10):\penalty0 2361--2379, 2019.

\bibitem[Cheng et~al.(2020)Cheng, Wang, Guan, and Yang]{Cheng2020CSPN++}
Xinjing Cheng, Peng Wang, Chenye Guan, and Ruigang Yang.
\newblock Cspn++: Learning context and resource aware convolutional spatial propagation networks for depth completion.
\newblock In \emph{AAAI}, pages 10615--10622, 2020.

\bibitem[Dai et~al.(2017)Dai, Qi, Xiong, Li, Zhang, Hu, and Wei]{dai2017deformable}
Jifeng Dai, Haozhi Qi, Yuwen Xiong, Yi Li, Guodong Zhang, Han Hu, and Yichen Wei.
\newblock Deformable convolutional networks.
\newblock In \emph{ICCV}, pages 764--773, 2017.

\bibitem[Dosovitskiy et~al.(2017)Dosovitskiy, Ros, Codevilla, Lopez, and Koltun]{dosovitskiy2017carla}
Alexey Dosovitskiy, German Ros, Felipe Codevilla, Antonio Lopez, and Vladlen Koltun.
\newblock Carla: An open urban driving simulator.
\newblock In \emph{CoRL}, pages 1--16. PMLR, 2017.

\bibitem[Eldesokey et~al.(2020)Eldesokey, Felsberg, and Khan]{2020Confidence}
Abdelrahman Eldesokey, Michael Felsberg, and Fahad~Shahbaz Khan.
\newblock Confidence propagation through cnns for guided sparse depth regression.
\newblock \emph{IEEE Transactions on Pattern Analysis and Machine Intelligence}, 42\penalty0 (10):\penalty0 2423--2436, 2020.

\bibitem[Gallego et~al.(2017)Gallego, Lund, Mueggler, Rebecq, Delbruck, and Scaramuzza]{gallego2017event}
Guillermo Gallego, Jon~EA Lund, Elias Mueggler, Henri Rebecq, Tobi Delbruck, and Davide Scaramuzza.
\newblock Event-based, 6-dof camera tracking from photometric depth maps.
\newblock \emph{IEEE Transactions on Pattern Analysis and Machine Intelligence}, 40\penalty0 (10):\penalty0 2402--2412, 2017.

\bibitem[Gallego et~al.(2018)Gallego, Rebecq, and Scaramuzza]{gallego2018unifying}
Guillermo Gallego, Henri Rebecq, and Davide Scaramuzza.
\newblock A unifying contrast maximization framework for event cameras, with applications to motion, depth, and optical flow estimation.
\newblock In \emph{CVPR}, pages 3867--3876, 2018.

\bibitem[Gallego et~al.(2020)Gallego, Delbr{\"u}ck, Orchard, Bartolozzi, Taba, Censi, Leutenegger, Davison, Conradt, Daniilidis, et~al.]{gallego2020event}
Guillermo Gallego, Tobi Delbr{\"u}ck, Garrick Orchard, Chiara Bartolozzi, Brian Taba, Andrea Censi, Stefan Leutenegger, Andrew~J Davison, J{\"o}rg Conradt, Kostas Daniilidis, et~al.
\newblock Event-based vision: A survey.
\newblock \emph{IEEE Transactions on Pattern Analysis and Machine Intelligence}, 44\penalty0 (1):\penalty0 154--180, 2020.

\bibitem[Gehrig and Scaramuzza(2024)]{gehrig2024low}
Daniel Gehrig and Davide Scaramuzza.
\newblock Low-latency automotive vision with event cameras.
\newblock \emph{Nature}, 629\penalty0 (8014):\penalty0 1034--1040, 2024.

\bibitem[Gehrig et~al.(2020)Gehrig, Gehrig, Hidalgo-Carri{\'o}, and Scaramuzza]{gehrig2020video}
Daniel Gehrig, Mathias Gehrig, Javier Hidalgo-Carri{\'o}, and Davide Scaramuzza.
\newblock Video to events: Recycling video datasets for event cameras.
\newblock In \emph{CVPR}, pages 3586--3595, 2020.

\bibitem[Gehrig et~al.(2021)Gehrig, R{\"u}egg, Gehrig, Hidalgo-Carri{\'o}, and Scaramuzza]{gehrig2021combining}
Daniel Gehrig, Michelle R{\"u}egg, Mathias Gehrig, Javier Hidalgo-Carri{\'o}, and Davide Scaramuzza.
\newblock Combining events and frames using recurrent asynchronous multimodal networks for monocular depth prediction.
\newblock \emph{IEEE Robotics and Automation Letters}, 6\penalty0 (2):\penalty0 2822--2829, 2021.

\bibitem[Geiger et~al.(2012)Geiger, Lenz, and Urtasun]{geiger2012we}
Andreas Geiger, Philip Lenz, and Raquel Urtasun.
\newblock Are we ready for autonomous driving? the kitti vision benchmark suite.
\newblock In \emph{CVPR}, pages 3354--3361, 2012.

\bibitem[Ghosh and Gallego(2022)]{ghosh2022event}
Suman Ghosh and Guillermo Gallego.
\newblock Event-based stereo depth estimation from ego-motion using ray density fusion.
\newblock \emph{arXiv preprint arXiv:2210.08927}, 2022.

\bibitem[Gregorek and Nalpantidis(2024)]{gregorek2024steeredmarigold}
Jakub Gregorek and Lazaros Nalpantidis.
\newblock Steeredmarigold: Steering diffusion towards depth completion of largely incomplete depth maps.
\newblock \emph{arXiv preprint arXiv:2409.10202}, 2024.

\bibitem[Hidalgo-Carri{\'o} et~al.(2020)Hidalgo-Carri{\'o}, Gehrig, and Scaramuzza]{hidalgo2020learning}
Javier Hidalgo-Carri{\'o}, Daniel Gehrig, and Davide Scaramuzza.
\newblock Learning monocular dense depth from events.
\newblock In \emph{3DV}, pages 534--542. IEEE, 2020.

\bibitem[Hitzges et~al.(2025)Hitzges, Ghosh, and Gallego]{hitzges2025derd}
Diego de~Oliveira Hitzges, Suman Ghosh, and Guillermo Gallego.
\newblock Derd-net: Learning depth from event-based ray densities.
\newblock \emph{arXiv preprint arXiv:2504.15863}, 2025.

\bibitem[Hu et~al.(2018)Hu, Shen, and Sun]{hu2018squeeze}
Jie Hu, Li Shen, and Gang Sun.
\newblock Squeeze-and-excitation networks.
\newblock In \emph{CVPR}, pages 7132--7141, 2018.

\bibitem[Hu et~al.(2021)Hu, Wang, Li, Ning, Fan, and Gong]{hu2020PENet}
Mu Hu, Shuling Wang, Bin Li, Shiyu Ning, Li Fan, and Xiaojin Gong.
\newblock Penet: Towards precise and efficient image guided depth completion.
\newblock In \emph{ICRA}, 2021.

\bibitem[Jaderberg et~al.(2015)Jaderberg, Simonyan, Zisserman, et~al.]{jaderberg2015spatial}
Max Jaderberg, Karen Simonyan, Andrew Zisserman, et~al.
\newblock Spatial transformer networks.
\newblock In \emph{NeurIPS}, 2015.

\bibitem[Jeon and Kim(2017)]{jeon2017active}
Yunho Jeon and Junmo Kim.
\newblock Active convolution: Learning the shape of convolution for image classification.
\newblock In \emph{CVPR}, pages 4201--4209, 2017.

\bibitem[Jia et~al.(2016)Jia, De~Brabandere, Tuytelaars, and Gool]{jia2016dynamic}
Xu Jia, Bert De~Brabandere, Tinne Tuytelaars, and Luc~V Gool.
\newblock Dynamic filter networks.
\newblock In \emph{NeurIPS}, 2016.

\bibitem[Kipf and Welling(2016)]{kipf2016semi}
Thomas~N Kipf and Max Welling.
\newblock Semi-supervised classification with graph convolutional networks.
\newblock \emph{arXiv preprint arXiv:1609.02907}, 2016.

\bibitem[Ku et~al.(2018)Ku, Harakeh, and Waslander]{ku2018ipbasic}
Jason Ku, Ali Harakeh, and Steven~L Waslander.
\newblock In defense of classical image processing: Fast depth completion on the cpu.
\newblock In \emph{CRV}, pages 16--22. IEEE, 2018.

\bibitem[Liang et~al.(2025)Liang, Hu, Shao, and Fu]{liang2025distilling}
Yingping Liang, Yutao Hu, Wenqi Shao, and Ying Fu.
\newblock Distilling monocular foundation model for fine-grained depth completion.
\newblock \emph{arXiv preprint arXiv:2503.16970}, 2025.

\bibitem[Lin et~al.(2022)Lin, Cheng, Zhong, Zhou, and Yang]{lin2022dynamic}
Yuankai Lin, Tao Cheng, Qi Zhong, Wending Zhou, and Hua Yang.
\newblock Dynamic spatial propagation network for depth completion.
\newblock In \emph{AAAI}, pages 1638--1646, 2022.

\bibitem[Lin et~al.(2023)Lin, Yang, Cheng, Zhou, and Yin]{lin2023dyspn}
Yuankai Lin, Hua Yang, Tao Cheng, Wending Zhou, and Zhouping Yin.
\newblock Dyspn: Learning dynamic affinity for image-guided depth completion.
\newblock \emph{IEEE Transactions on Circuits and Systems for Video Technology}, pages 1--1, 2023.

\bibitem[Liu et~al.(2021)Liu, Song, Lyu, Diao, Wang, Liu, and Zhang]{liu2021fcfr}
Lina Liu, Xibin Song, Xiaoyang Lyu, Junwei Diao, Mengmeng Wang, Yong Liu, and Liangjun Zhang.
\newblock Fcfr-net: Feature fusion based coarse-to-fine residual learning for depth completion.
\newblock In \emph{AAAI}, pages 2136--2144, 2021.

\bibitem[Liu et~al.(2024)Liu, Li, Shi, Fan, Tian, and Zhao]{liu2024event}
Xu Liu, Jianing Li, Jinqiao Shi, Xiaopeng Fan, Yonghong Tian, and Debin Zhao.
\newblock Event-based monocular depth estimation with recurrent transformers.
\newblock \emph{IEEE Transactions on Circuits and Systems for Video Technology}, 2024.

\bibitem[Loshchilov and Hutter(2017)]{loshchilov2017decoupled}
Ilya Loshchilov and Frank Hutter.
\newblock Decoupled weight decay regularization.
\newblock \emph{arXiv preprint arXiv:1711.05101}, 2017.

\bibitem[Lu et~al.(2020)Lu, Barnes, Anwar, and Zheng]{2020FromLu}
Kaiyue Lu, Nick Barnes, Saeed Anwar, and Liang Zheng.
\newblock From depth what can you see? depth completion via auxiliary image reconstruction.
\newblock In \emph{CVPR}, pages 11306--11315, 2020.

\bibitem[Ma and Karaman(2018)]{ma2018sparse}
Fangchang Ma and Sertac Karaman.
\newblock Sparse-to-dense: Depth prediction from sparse depth samples and a single image.
\newblock In \emph{ICRA}, pages 4796--4803. IEEE, 2018.

\bibitem[Muglikar et~al.(2021)Muglikar, Moeys, and Scaramuzza]{muglikar2021event}
Manasi Muglikar, Diederik~Paul Moeys, and Davide Scaramuzza.
\newblock Event guided depth sensing.
\newblock In \emph{3DV}, pages 385--393. IEEE, 2021.

\bibitem[Pan et~al.(2024)Pan, Cao, and Wang]{pan2024srfnet}
Tianbo Pan, Zidong Cao, and Lin Wang.
\newblock Srfnet: Monocular depth estimation with fine-grained structure via spatial reliability-oriented fusion of frames and events.
\newblock In \emph{ICRA}, pages 10695--10702. IEEE, 2024.

\bibitem[Park et~al.(2020)Park, Joo, Hu, Liu, and Kweon]{park2020nonlocal}
Jinsun Park, Kyungdon Joo, Zhe Hu, Chi-Kuei Liu, and In~So Kweon.
\newblock Non-local spatial propagation network for depth completion.
\newblock In \emph{ECCV}, 2020.

\bibitem[Park and Jeon(2024)]{park2024simple}
Jin-Hwi Park and Hae-Gon Jeon.
\newblock A simple yet universal framework for depth completion.
\newblock In \emph{NeurIPS}, 2024.

\bibitem[Park et~al.(2024)Park, Jeong, Lee, and Jeon]{park2024depth}
Jin-Hwi Park, Chanhwi Jeong, Junoh Lee, and Hae-Gon Jeon.
\newblock Depth prompting for sensor-agnostic depth estimation.
\newblock In \emph{CVPR}, pages 9859--9869, 2024.

\bibitem[Rebecq et~al.(2018)Rebecq, Gallego, Mueggler, and Scaramuzza]{rebecq2018emvs}
Henri Rebecq, Guillermo Gallego, Elias Mueggler, and Davide Scaramuzza.
\newblock Emvs: Event-based multi-view stereo—3d reconstruction with an event camera in real-time.
\newblock \emph{International Journal of Computer Vision}, 126\penalty0 (12):\penalty0 1394--1414, 2018.

\bibitem[Shao et~al.(2023)Shao, Pei, Chen, Wu, and Li]{shao2023nddepth}
Shuwei Shao, Zhongcai Pei, Weihai Chen, Xingming Wu, and Zhengguo Li.
\newblock Nddepth: Normal-distance assisted monocular depth estimation.
\newblock In \emph{ICCV}, pages 7931--7940, 2023.

\bibitem[Shiba et~al.(2024)Shiba, Klose, Aoki, and Gallego]{shiba2024secrets}
Shintaro Shiba, Yannick Klose, Yoshimitsu Aoki, and Guillermo Gallego.
\newblock Secrets of event-based optical flow, depth and ego-motion estimation by contrast maximization.
\newblock \emph{IEEE Transactions on Pattern Analysis and Machine Intelligence}, 2024.

\bibitem[Silberman et~al.(2012)Silberman, Hoiem, Kohli, and Fergus]{silberman2012indoor}
Nathan Silberman, Derek Hoiem, Pushmeet Kohli, and Rob Fergus.
\newblock Indoor segmentation and support inference from rgbd images.
\newblock In \emph{ECCV}, pages 746--760. Springer, 2012.

\bibitem[Smith and Topin(2019)]{smith2019super}
Leslie~N Smith and Nicholay Topin.
\newblock Super-convergence: Very fast training of neural networks using large learning rates.
\newblock In \emph{Artificial Intelligence and Machine Learning for Multi-domain Operations Applications}, pages 369--386. SPIE, 2019.

\bibitem[Song et~al.(2015)Song, Lichtenberg, and Xiao]{song2015sun}
Shuran Song, Samuel~P Lichtenberg, and Jianxiong Xiao.
\newblock Sun rgb-d: A rgb-d scene understanding benchmark suite.
\newblock In \emph{CVPR}, pages 567--576, 2015.

\bibitem[Song et~al.(2020)Song, Dai, Zhou, Liu, Li, Li, and Yang]{song2020channel}
Xibin Song, Yuchao Dai, Dingfu Zhou, Liu Liu, Wei Li, Hongdong Li, and Ruigang Yang.
\newblock Channel attention based iterative residual learning for depth map super-resolution.
\newblock In \emph{CVPR}, pages 5631--5640, 2020.

\bibitem[Tang et~al.(2020)Tang, Tian, Feng, Li, and Tan]{tang2020learning}
Jie Tang, Fei-Peng Tian, Wei Feng, Jian Li, and Ping Tan.
\newblock Learning guided convolutional network for depth completion.
\newblock \emph{IEEE Transactions on Image Processing}, 30:\penalty0 1116--1129, 2020.

\bibitem[Tang et~al.(2024)Tang, Tian, An, Li, and Tan]{tang2024bilateral}
Jie Tang, Fei-Peng Tian, Boshi An, Jian Li, and Ping Tan.
\newblock Bilateral propagation network for depth completion.
\newblock In \emph{CVPR}, pages 9763--9772, 2024.

\bibitem[Uhrig et~al.(2017)Uhrig, Schneider, Schneider, Franke, Brox, and Geiger]{Uhrig2017THREEDV}
Jonas Uhrig, Nick Schneider, Lukas Schneider, Uwe Franke, Thomas Brox, and Andreas Geiger.
\newblock Sparsity invariant cnns.
\newblock In \emph{3DV}, pages 11--20, 2017.

\bibitem[Van~Gansbeke et~al.(2019)Van~Gansbeke, Neven, De~Brabandere, and Van~Gool]{vangansbeke2019}
Wouter Van~Gansbeke, Davy Neven, Bert De~Brabandere, and Luc Van~Gool.
\newblock Sparse and noisy lidar completion with rgb guidance and uncertainty.
\newblock In \emph{MVA}, pages 1--6, 2019.

\bibitem[Veli{\v{c}}kovi{\'c} et~al.(2017)Veli{\v{c}}kovi{\'c}, Cucurull, Casanova, Romero, Lio, and Bengio]{velivckovic2017graph}
Petar Veli{\v{c}}kovi{\'c}, Guillem Cucurull, Arantxa Casanova, Adriana Romero, Pietro Lio, and Yoshua Bengio.
\newblock Graph attention networks.
\newblock \emph{arXiv preprint arXiv:1710.10903}, 2017.

\bibitem[Viola et~al.(2024)Viola, Qu, Metzger, Ke, Becker, Schindler, and Obukhov]{viola2024marigold}
Massimiliano Viola, Kevin Qu, Nando Metzger, Bingxin Ke, Alexander Becker, Konrad Schindler, and Anton Obukhov.
\newblock Marigold-dc: Zero-shot monocular depth completion with guided diffusion.
\newblock \emph{arXiv preprint arXiv:2412.13389}, 2024.

\bibitem[Wang et~al.(2025)Wang, Yan, Fan, Li, and Yang]{wang2025learning}
Kun Wang, Zhiqiang Yan, Junkai Fan, Jun Li, and Jian Yang.
\newblock Learning inverse laplacian pyramid for progressive depth completion.
\newblock \emph{arXiv preprint arXiv:2502.07289}, 2025.

\bibitem[Wang et~al.(2023)Wang, Li, Zhang, Liu, Gao, and Dai]{wang2023lrru}
Yufei Wang, Bo Li, Ge Zhang, Qi Liu, Tao Gao, and Yuchao Dai.
\newblock Lrru: Long-short range recurrent updating networks for depth completion.
\newblock In \emph{ICCV}, pages 9422--9432, 2023.

\bibitem[Wang et~al.(2024)Wang, Zhang, Wang, Li, Liu, Hui, and Dai]{wang2024improving}
Yufei Wang, Ge Zhang, Shaoqian Wang, Bo Li, Qi Liu, Le Hui, and Yuchao Dai.
\newblock Improving depth completion via depth feature upsampling.
\newblock In \emph{CVPR}, pages 21104--21113, 2024.

\bibitem[Wei et~al.(2024)Wei, Jiao, Hu, Yu, Xie, Wu, Zhu, Liu, Wang, and Liu]{wei2024fusionportablev2}
Hexiang Wei, Jianhao Jiao, Xiangcheng Hu, Jingwen Yu, Xupeng Xie, Jin Wu, Yilong Zhu, Yuxuan Liu, Lujia Wang, and Ming Liu.
\newblock Fusionportablev2: A unified multi-sensor dataset for generalized slam across diverse platforms and scalable environments.
\newblock \emph{The International Journal of Robotics Research}, page 02783649241303525, 2024.

\bibitem[Xu et~al.(2020)Xu, Yin, and Yao]{xu2020deformable}
Zheyuan Xu, Hongche Yin, and Jian Yao.
\newblock Deformable spatial propagation networks for depth completion.
\newblock In \emph{ICIP}, pages 913--917. IEEE, 2020.

\bibitem[Yan et~al.(2022{\natexlab{a}})Yan, Wang, Li, Zhang, Li, Li, and Yang]{yan2022learning}
Zhiqiang Yan, Kun Wang, Xiang Li, Zhenyu Zhang, Guangyu Li, Jun Li, and Jian Yang.
\newblock Learning complementary correlations for depth super-resolution with incomplete data in real world.
\newblock \emph{IEEE Transactions on Neural Networks and Learning Systems}, 2022{\natexlab{a}}.

\bibitem[Yan et~al.(2022{\natexlab{b}})Yan, Wang, Li, Zhang, Li, and Yang]{yan2022rignet}
Zhiqiang Yan, Kun Wang, Xiang Li, Zhenyu Zhang, Jun Li, and Jian Yang.
\newblock Rignet: Repetitive image guided network for depth completion.
\newblock In \emph{ECCV}, pages 214--230, 2022{\natexlab{b}}.

\bibitem[Yan et~al.(2023{\natexlab{a}})Yan, Li, Zhang, Li, and Yang]{yan2023rignet++}
Zhiqiang Yan, Xiang Li, Zhenyu Zhang, Jun Li, and Jian Yang.
\newblock Rignet++: Efficient repetitive image guided network for depth completion.
\newblock \emph{arXiv preprint arXiv:2309.00655}, 2023{\natexlab{a}}.

\bibitem[Yan et~al.(2023{\natexlab{b}})Yan, Wang, Li, Zhang, Li, and Yang]{yan2023desnet}
Zhiqiang Yan, Kun Wang, Xiang Li, Zhenyu Zhang, Jun Li, and Jian Yang.
\newblock Desnet: Decomposed scale-consistent network for unsupervised depth completion.
\newblock In \emph{AAAI}, pages 3109--3117, 2023{\natexlab{b}}.

\bibitem[Yan et~al.(2023{\natexlab{c}})Yan, Zheng, Wang, Li, Zhang, Chen, Li, and Yang]{yan2023learnable}
Zhiqiang Yan, Yupeng Zheng, Kun Wang, Xiang Li, Zhenyu Zhang, Shuo Chen, Jun Li, and Jian Yang.
\newblock Learnable differencing center for nighttime depth perception.
\newblock \emph{arXiv preprint arXiv:2306.14538}, 2023{\natexlab{c}}.

\bibitem[Yan et~al.(2024{\natexlab{a}})Yan, Lin, Wang, Zheng, Wang, Zhang, Li, and Yang]{yan2024tri}
Zhiqiang Yan, Yuankai Lin, Kun Wang, Yupeng Zheng, Yufei Wang, Zhenyu Zhang, Jun Li, and Jian Yang.
\newblock Tri-perspective view decomposition for geometry-aware depth completion.
\newblock In \emph{CVPR}, pages 4874--4884, 2024{\natexlab{a}}.

\bibitem[Yan et~al.(2024{\natexlab{b}})Yan, Wang, Wang, Li, and Yang]{yan2024completion}
Zhiqiang Yan, Zhengxue Wang, Kun Wang, Jun Li, and Jian Yang.
\newblock Completion as enhancement: A degradation-aware selective image guided network for depth completion.
\newblock \emph{arXiv preprint arXiv:2412.19225}, 2024{\natexlab{b}}.

\bibitem[Yu et~al.(2023)Yu, Sheng, Zhou, Luo, Cao, Gu, Zhang, and Shen]{yu2023aggregating}
Zhu Yu, Zehua Sheng, Zili Zhou, Lun Luo, Si-Yuan Cao, Hong Gu, Huaqi Zhang, and Hui-Liang Shen.
\newblock Aggregating feature point cloud for depth completion.
\newblock In \emph{ICCV}, pages 8732--8743, 2023.

\bibitem[Zhang et~al.(2024)Zhang, Han, Sobeih, Han, and Dancey]{zhang2024novel}
Xin Zhang, Liangxiu Han, Tam Sobeih, Lianghao Han, and Darren Dancey.
\newblock A novel spike transformer network for depth estimation from event cameras via cross-modality knowledge distillation.
\newblock \emph{arXiv preprint arXiv:2404.17335}, 2024.

\bibitem[Zhang et~al.(2023)Zhang, Guo, Poggi, Zhu, Huang, and Mattoccia]{zhang2023cf}
Youmin Zhang, Xianda Guo, Matteo Poggi, Zheng Zhu, Guan Huang, and Stefano Mattoccia.
\newblock Completionformer: Depth completion with convolutions and vision transformers.
\newblock In \emph{CVPR}, pages 18527--18536, 2023.

\bibitem[Zhao et~al.(2021)Zhao, Gong, Fu, and Tao]{zhao2021adaptive}
Shanshan Zhao, Mingming Gong, Huan Fu, and Dacheng Tao.
\newblock Adaptive context-aware multi-modal network for depth completion.
\newblock \emph{IEEE Transactions on Image Processing}, 2021.

\bibitem[Zhou et~al.(2023)Zhou, Yan, Liao, Lin, Huang, Zhao, Cui, and Li]{zhou2023bev}
Wending Zhou, Xu Yan, Yinghong Liao, Yuankai Lin, Jin Huang, Gangming Zhao, Shuguang Cui, and Zhen Li.
\newblock Bev@ dc: Bird's-eye view assisted training for depth completion.
\newblock In \emph{CVPR}, pages 9233--9242, 2023.

\bibitem[Zhu et~al.(2019{\natexlab{a}})Zhu, Yuan, Chaney, and Daniilidis]{zhu2019unsupervised}
Alex~Zihao Zhu, Liangzhe Yuan, Kenneth Chaney, and Kostas Daniilidis.
\newblock Unsupervised event-based learning of optical flow, depth, and egomotion.
\newblock In \emph{CVPR}, pages 989--997, 2019{\natexlab{a}}.

\bibitem[Zhu et~al.(2023)Zhu, Liu, Jiang, Wen, Zhang, Li, and Liu]{zhu2023self}
Junyu Zhu, Lina Liu, Bofeng Jiang, Feng Wen, Hongbo Zhang, Wanlong Li, and Yong Liu.
\newblock Self-supervised event-based monocular depth estimation using cross-modal consistency.
\newblock In \emph{IROS}, pages 7704--7710. IEEE, 2023.

\bibitem[Zhu et~al.(2019{\natexlab{b}})Zhu, Hu, Lin, and Dai]{zhu2019deformable}
Xizhou Zhu, Han Hu, Stephen Lin, and Jifeng Dai.
\newblock Deformable convnets v2: More deformable, better results.
\newblock In \emph{CVPR}, pages 9308--9316, 2019{\natexlab{b}}.

\bibitem[Zihao~Zhu et~al.(2017)Zihao~Zhu, Atanasov, and Daniilidis]{zihao2017event}
Alex Zihao~Zhu, Nikolay Atanasov, and Kostas Daniilidis.
\newblock Event-based visual inertial odometry.
\newblock In \emph{CVPR}, pages 5391--5399, 2017.

\bibitem[Zuo and Deng(2024)]{zuo2024ogni}
Yiming Zuo and Jia Deng.
\newblock Ogni-dc: Robust depth completion with optimization-guided neural iterations.
\newblock In \emph{ECCV}, pages 78--95. Springer, 2024.

\end{thebibliography}
}

\end{document}